\pdfoutput=1

\documentclass[11pt]{article}

\usepackage{ACL2023}
\usepackage{CJKutf8} 
\usepackage{times}
\usepackage{multirow}
\usepackage{latexsym}
\usepackage{hhline}

\usepackage[T1]{fontenc}

\usepackage[utf8]{inputenc}
\usepackage{microtype}

\usepackage{inconsolata}
\usepackage{graphicx}

\title{Chinese MentalBERT: Domain-Adaptive Pre-training on Social Media for Chinese Mental Health Text Analysis}

\author{Wei Zhai\textsuperscript{1}, Hongzhi Qi\textsuperscript{1}, Qing Zhao\textsuperscript{1}, Jianqiang Li\textsuperscript{1}, Ziqi Wang\textsuperscript{1}, Han Wang\textsuperscript{1}, \\
\textbf{Bing Xiang Yang}\textsuperscript{2}, \textbf{Guanghui Fu}\textsuperscript{3}\thanks{*Corresponding author: guanghui.fu@inria.fr}  \\
\textsuperscript{1}School of Software Engineering, Beijing University of Technology, Beijing, China \\
\textsuperscript{2}School of Nursing, Wuhan University, Wuhan, China \\
\textsuperscript{3}Sorbonne Université, ICM, CNRS, Inria, Inserm, AP-HP, Hôpital de la Pitié-Salpêtrière, \\Paris, France}

\begin{document}
\pagestyle{plain}

\maketitle

\begin{abstract}
In the current environment, psychological issues are prevalent and widespread, with social media serving as a key outlet for individuals to share their feelings. This results in the generation of vast quantities of data daily, where negative emotions have the potential to precipitate crisis situations. There is a recognized need for models capable of efficient analysis. While pre-trained language models have demonstrated their effectiveness broadly, there's a noticeable gap in pre-trained models tailored for specialized domains like psychology. To address this, we have collected a huge dataset from Chinese social media platforms and enriched it with publicly available datasets to create a comprehensive database encompassing 3.36 million text entries. To enhance the model's applicability to psychological text analysis, we integrated psychological lexicons into the pre-training masking mechanism. Building on an existing Chinese language model, we performed adaptive training to develop a model specialized for the psychological domain. We evaluated our model’s performance across six public datasets, where it demonstrated improvements compared to eight other models. Additionally, in the qualitative comparison experiment, our model provided psychologically relevant predictions given the masked sentences. Due to concerns regarding data privacy, the dataset will not be made publicly available. However, we have made the pre-trained models and codes publicly accessible to the community via: \url{https://github.com/zwzzzQAQ/Chinese-MentalBERT}.
\end{abstract}

\section{Introduction}
Mental illnesses, particularly depression, impose a considerable strain on global societies. The World Health Organization reports that approximately 3.8\% of the global population suffers from depression~\cite{world2023depressive}. 
Notably, the incidence of depression in China accounts for as high as 6.9\% of the prevalence~\cite{huang2019prevalence}. 
Individuals experiencing emotional distress often resort to passive coping mechanisms and seldom seek professional help~\cite{rusch2005mental}. Traditional channels for emotional crisis intervention, such as hotlines and psychological clinics, are not designed to proactively identify individuals facing emotional challenges~\cite{world2014preventing}. Moreover, the resources for such interventions are frequently inadequate. The stigma associated with mental illness has led many to use social networks as a primary outlet for expressing their emotional struggles~\cite{primack2017social}. Platforms like X (Twitter), and Sina Weibo in China serve as venues for individuals to share their feelings and opinions in real time, with posts often providing immediate insights into one's daily experiences and emotional states~\cite{de2013predicting}. Within specific topics or hashtags on social media, there is a pronounced focus on the expression of negative emotions, with some users displaying evident suicidal tendencies~\cite{robinson2016social}. This situation underscores the critical necessity to develop tools aimed at enabling the early detection of such distress signals and implementing timely intervention strategies~\cite{coppersmith2018natural}. 

Pre-trained language models (PLMs), such as BERT~\cite{devlin2018bert}, have demonstrated remarkable success across a variety of language tasks and have seen extensive application~\cite{koroteev2021bert}. Recently, the development of large language model (LLM) technology has garnered global interest, notably within the psychology sector, where a plenty of exploratory applications have been initiated~\cite{he2023towards}. However, according to research by ~\citet{qi2023evaluating}, in comparison to supervised learning methods, LLMs are yet to fully addressed the complexity of psychological tasks. Thus, the development of PLMs specifically targeting for supervised learning still crucial, particularly for the specific domains.

To address the lack of large-scale PLMs tailored for specific applications in Chinese community mental health, we introduce the Chinese MentalBERT, a pre-trained language model specifically designed for psychological tasks. To the best of our knowledge, it is the first PLM developed specifically for the mental health field in Chinese. 
In this study, we employ a domain-adaptive pre-training model~\cite{chinese_bert_cui2021pre}, and introduce a novel lexicon guided masking mechanism strategy based on the Chinese depression lexicon. We conduct four Chinese mental health datasets from social media and public dataset, including over 3.36 million data items for domain-adaptive pre-training.
This lexion guided masking mechanism strategically biases the learning process towards vocabulary crucial for the intended application, enhancing the model's relevance and effectiveness in its target domain. 
We evaluated the performance of our model on six mental health-related public datasets, including: two cognitive distortion classifications, one suicide risk classification, and three sentiment analysis tasks. The results demonstrate that our model outperforms eight comparison methods across six public datasets. Due to privacy issues, we cannot open the data, we have made all the training code and models publicly available to support research in Chinese mental health.

\section{Related work}
Social media has been used to identify signs of needing medical or psychological support~\cite{keles2020systematic}. Recent studies show that mental health analyses on social media are using NLP technologies to capture users' mental health states~\cite{calvo2017natural}.
With the advent of BERT~\cite{devlin2018bert}, studies have been carried out on using this technology to assess suicidal tendencies and identify depressive tendencies~\cite{wang2019assessing,ambalavanan2019using,matero2019suicide}.
~\citet{yang2022characteristics} leveraged knowledge graph method to screen high-suicide risk comments within online forums and explored various attributes such as time, content, and suicidal behavior patterns by analyzing these comments. 
~\citet{fu2021distant} proposed a distant supervision method to develop an automated method capable of categorizing users into high or low suicide risk categories based on their social media comments. This model serves as an early warning system to aid volunteers in preventing potential suicides among social media users.

On the other hand, many studies focus on domain-adaptive pretraining in specific fields to pursue better domain performance. 
~\citet{chalkidis2020legal} systematically examined methods for adapting BERT to the legal field. They gathered 12 GB of varied English legal text from public sources and achieved improved performance compared to the baseline on three end-tasks.
~\citet{lee2020biobert} pre-trained the BERT model with a domain-specific regimen on extensive biomedical corpora, leading to improved performance compared to the original BERT across various text mining tasks in the biomedical field.

The two most closely related works to our study are the research in the domain of mental health data analysis.
~\citet{ji2021mentalbert} implemented BERT within the mental health sector by creating a targeted dataset from Reddit, resulting in the development of a model known as MentalBERT. 
Subsequently,~\citet{aragon2023disorbert} introduced a dual-domain adaptation process for language models pretraining, which involves firstly adapting the model to the social media text and then to the mental health domain. Throughout both stages, the integration of lexical resources played a critical role in directing the language model's masking procedure, thereby ensuring a heightened focus on vocabulary associated with mental disorders. 
Currently, there is no PLM customized for the Chinese mental health domain, and unique challenges exist within the Chinese field. 
To bridge this gap, we collected around 4 million data from social media for domain adaptive pre-training in the Chinese mental health domain. The guided masking mechanism based on the domain lexicon can help the model focus on the domain-related context.
\begin{figure*}[!htbp]
\centering
\includegraphics[width=0.8\linewidth]{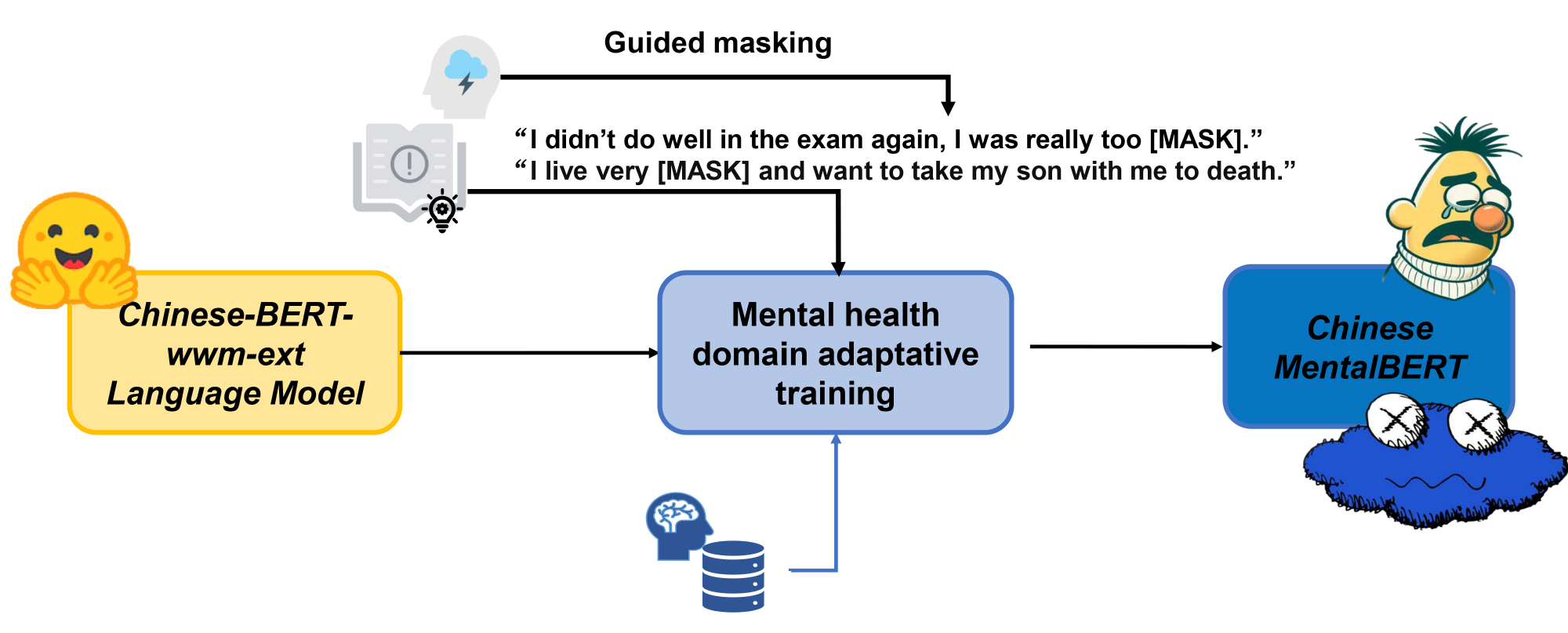} 
\caption{Overview of the domain adaptive pretraining process. The process initiates with the basic pretrained language model (Chinese-BERT-wwm-ext), followed by further pretraining with 3.36 millions mental health posts/comments sourced from social media. The pretraining phase integrates the knowledge from depression lexicon to guide the masking process.}
\label{fig:diagram} 
\end{figure*}

\section{Pretraining Corpus}
\begin{itemize}
    \item Comment from ``Zoufan'' Weibo treehole~\cite{zoufan}: ``Zoufan'' Weibo accounts are often likened to digital confession booths or ``tree holes'', where users predominantly express negative emotions. Some users have even expressed suicidal intentions on it. By 2020, we had collected roughly 2.34 million comments from 351,069 users. However, from 2021 onwards, data collection has been halted due to platform restrictions.
    \item Weibo Depression ``Chaohua'' (super topic on Weibo)~\cite{chaohua}: it functions as a specialized sub-forum intended for communication and information exchange among individuals dealing with depression or those interested in learning more about it. This digital space enables users to share personal experiences, discuss treatment options, provide mental health resources, and offer insights and understandings related to depression. In total, we collected 504,072 posts from 69,102 users.
    \item Sina Weibo Depression Dataset (SWDD)~\cite{dataset_cai2023depression}\footnote{\url{https://github.com/ethan-nicholas-tsai/DepressionDetection}}: this dataset represents a comprehensive collection of depression-related data from Sina Weibo, including complete user posting histories up to January 1, 2021. It encompasses user profiles and their entire post history, focusing exclusively on depression-related content. Our analysis leveraged only depression-related user and posts and included data from 3,711 users and a total of 785,689 posts.
    \item Weibo User Depression Detection Dataset (WU3D)~\cite{dataset_wang2020multitask}\footnote{\url{https://github.com/aidenwang9867/Weibo-User-Depression-Detection-Dataset}}: WU3D contains enriched information fields, including posts, the posting time, posted pictures, the user gender, etc. This dataset is labeled and further reviewed by professionals. We  use the depressed user's posts data including 10,325 users and 408,797 posts.
\end{itemize}

\begin{table}
\centering
\begin{tabular}{|c|c|c|} 
\hline
Dataset                 & Users   & posts     \\ 
\hline
``Zoufan'' Weibo treehole & 351,069 & 2,346,879  \\
Depression ``Chaohua''    & 69,102  & 504,072    \\
SWDD                    & 3,711   & 785,689    \\
WU3D                    & 10,325  & 408,797    \\ 
\hline
In total                & 434,207 & 4,045,437  \\
\hline
\end{tabular}
\caption{Distribution of pretraining corpus datasets, including the number of users and posts in each dataset.}
\label{tab:corpus_stats}
\end{table}

Ultimately, our corpus comprised over 4.04 million posts/comments from 434,207 users. After thoroughly cleaning the dataset by removing excessively short and meaningless sentences, we compiled 3,360,273 data. This cleaned dataset served as the pre-training set for our model.

\section{Methods}
\subsection{Basic pre-training language model: Chinese-BERT-wwm-ext} \label{sec:methods:pretrained}
The proposed Chinese MentalBERT is a domain adaptive pretrained version of Chinese-BERT-wwm-ext~\cite{chinese_bert_cui2021pre}. Domain-adaptive pretraining has been proven to be effective~\cite{domain_adaptative_gururangan2020don}. The continual pretraining can benifit for the targeted downstream domain~\cite{jin2021lifelong}. The original pretrained model acquires knowledge from general domains. Our adaptive pre-training process enhances the model's ability to perform tasks specifically within the Chinese comunity mental health domain.
Chinese-BERT-wwm-ext integrates a ``Whole Word Masking'' (WWM) strategy in its pre-training phase~\cite{chinese_bert_cui2021pre}. 
Unlike the original BERT~\cite{devlin2018bert}, which typically masks only a single English word, Chinese-BERT-wwm-ext masks to mask entire Chinese words. This approach is due to the lexical differences between Chinese and English: English uses single words to convey meanings, while Chinese relies on compound words composed of multiple characters for complete concepts. Masking single characters in Chinese leads to incomplete meanings, obstructing the model's learning of the language's structure.
The pretraining data is comprised of Chinese Wikipedia dump\footnote{\url{https://dumps.wikimedia.org/zhwiki/latest}}, alongside an extensive collection of additional data including encyclopedic content, news articles, and question-answering websites, encompassing a total of 5.4 billion words.

\subsection{Masking mechanism guided by depression lexicon}
As previously discussed, the masking mechanism plays a pivotal role in the pre-training of language models. The pretrained model in general domain such as BERT and Chinese-BERT-wwm-ext, typically employ random masking. This technique involves selecting random words within a sentence to be masked, challenging the model to predict these hidden words based on their context. Although random masking is a proven method for enhancing a model's contextual understanding, the development of knowledge-guided masking strategies represents an advanced step towards crafting domain-specific language models~\cite{tian2020skep, shamshiri2024text}.

To better tailor our model to the specific needs of the mental health domain, we implemented a guided masking strategy utilizing a depression lexicon. This approach begins by identifying whether the pre-training text contains lexicon words; if so, these words are masked for prediction training. Should the proportion of text occluded fall below 20\%, we augment the masked selection with additional, randomly chosen words. It's important to note the distinct strategies required for word guidance and masking in English versus Chinese texts. While masking in word-level suffices in English, Chinese requires word segmentation to mask compound words accurately, ensuring complete concepts are expressed and understood by the model.

Our research investigates a lexicon-guided masking mechanism. The depression lexicon is developed by~\citet{lexicon_li2020automatic}. This lexicon is derived from a labeled dataset of depression-related posts on Weibo, comprising 111,052 posts from 1,868 users, including both depression-related and non-depression-related content. The lexicon construction employs 80 seed words to establish semantic associations between these seeds and potential candidate words, forming a semantic association graph. The label propagation algorithm (LPA) is then applied to automatically assign labels to new words within this graph. This enriched dictionary serves as a input for machine learning algorithms, improving the performance to detect a test subject's depressive state. 

\section{Experimental Settings}
\subsection{Dataset}
The proposed pretrained model underwent evaluation on six public datasets in the mental health domain, including sentiment analysis, cognitive distortion identification and suicide detection. The distribution of experimental datasets can be seen in Table~\ref{tab:datasets}.
\begin{table*}
\centering
\begin{tabular}{ccccccc}
\hline
Dataset & $N_{train}$ & $N_{val}$ &$N_{test}$ & $C$ & $\overline{C}$ & $\overline{W}$\\ 
\hline
EWECT-usual & 27768 & 2000 & 5000 & 6 & 1 & 44.16\\
\hline
EWECT-epidemic & 8606 & 2000 & 3000 & 6 & 1 & 51.38\\
\hline
Waimai\_10k & 9589 & 1199 & 1199 & 2 & 1 & 25.04\\
\hline
Cognitive & 2180 & 545 & 682 & 12 & 1.2 & 41.59\\
\hline
C2D2 & 4500 & 1500 & 1500 & 8 & 1 & 29.68\\
\hline
Suicide & 800 & 199 & 250 & 2 & 1 & 47.79\\
\hline
\end{tabular}
\caption{Distribution of the experimental datasets. $N_{train}, N_{val}, N_{test}$ represent the number of items in the training, validation and test sets, respectively. $C$ represent the number of categories within the task, while $\overline{C}$ denotes the average number of categories across tasks per sample. And $\overline{W}$ average number of words per sample.}
\label{tab:datasets}    
\end{table*}

\subsubsection{Sentiment analysis tasks}
We conducted experiments on three sentiment analysis tasks, using two datasets from SMP2020-EWECT\footnote{\url{https://github.com/BrownSweater/BERT_SMP2020-EWECT}} for social media emotional classification and Waimai\_10k dataset\footnote{\url{https://github.com/SophonPlus/ChineseNlpCorpus/tree/master/datasets/waimai_10k}} for positive and negative sentiment classification on delivery platforms.
The SMP2020-EWECT includes two datasets: the first consists of posts randomly collected from Weibo social media platform on various topics; the second consists of COVID epidemic-related posts from Weibo. The objective of the two SMP2020-EWECT tasks are to identify the six categories of emotions in the posts: positive, angry, sad, fearful, surprised, and no emotion. 
The Waimai\_10k dataset is derived from user reviews collected on a food takeout platform, and the task is to identify positive and negative emotions.

\subsubsection{Cognitive distortion classification tasks}
In the cognitive distortion classification task, we conducted experiments on two datasets: ``Cognitive''~\cite{qi2023evaluating} and ``C2D2''~\cite{wang2023c2d2}. 
\begin{itemize}
    \item ``Cognitive'' dataset: is from Weibo, sourced from the ``Zoufan'' tree hole. The cognitive distortion task centers on the categories defined by ~\citet{burns1981feeling}. Data were obtained by crawling comments from the ``Zoufan'' blog on the Weibo social platform. Subsequently, a team of qualified psychologists was recruited to annotate the data. The experts define it as a multi-label classification task since each post may reflect multiple cognitive distortions. The classification labels in the cognitive distortion dataset include: all-or-nothing thinking, over-generalization, mental filter, disqualifying the positive, mind reading, the fortune teller error, magnification, emotional reasoning, should statements, labeling and mislabeling, blaming oneself and blaming others. Given that the data are publicly accessible, privacy concerns are not applicable. 
    \item ``C2D2'' dataset: it is a Chinese cognitive distortion dataset, containing 7,500 entries of cognitively distorted thoughts from daily life scenarios. The data annotation process is conducted through a collaborative effort between carefully selected, specially trained volunteers and domain experts. They treat it as a multi-class classification task because it simplifies the annotating process. The classification labels in the cognitive distortion dataset include: black-and-white thinking, emotional reasoning, fortune-telling, labeling, mindreading, overgeneralization, personalization, and non-distorted.
\end{itemize}

\subsubsection{Suicide intention classification task}
This dataset is also from Weibo, specifically collected from the ``Zoufan'' tree hole, as detailed in the study by~\citet{qi2023evaluating}.
The suicide risk task aims to differentiate between high and low suicide risk.  
For the suicide detection data, the dataset contained 645 records with low suicide risk and 601 records with high suicide risk.

\subsection{Implementation Details}
\subsubsection{Domain adaptive pre-training}
We conducted text preprocessing to eliminate irrelevant information, which included URLs, user tags (e.g., @username), topic tags (e.g., \verb|#topic#|), and we also removed special symbols, emoticons, and unstructured characters. Following this preprocessing, we concatenated all samples and segmented the entire corpus into equal-sized chunks, each consisting of 128 words.

The domain adaptive pretrained model is tasked with predicting these masked words in the training process. The words appeared in the depression lexion are preferentially masked within the sample. If the proportion of masked words in the original text less than the required 20\%, additional random words are incorporated into the mask to meet this threshold. 

The pretraining was performed on an NVIDIA Tesla V100 32GB SXM2 GPU. Building upon the foundational pretrained model (Chinese-BERT-wwm-ext), we continued training three epochs, utilizing a batch size of 128 and a learning rate of $5e^{-5}$.

\subsubsection{Finetuning for downstream tasks}
We followed the same pre-training text preprocessing step, removing extraneous information such as URLs, user tags, and topic tags to ensure the cleanliness and relevance of the dataset.
Both the supervised learning models and large language models are fine-tuned for downstream tasks.
\begin{itemize}
    \item Supervised learning models: These models were fine-tuned for 10 epochs on the training set for all these tasks. We employed a batch size of 16, utilized the Adam optimizer~\cite{kingma2014adam}, set the learning rate to $2e^{-5}$, and used cross-entropy as the loss function. The model was trained using an NVIDIA GeForce RTX 4090 24GB GPU.
    \item Large language model: The LLMs were fine-tuned for 5 epochs using the LoRA technique~\cite{hu2021lora} on the training set. LoRA is recognized as one of the most effective methods for fine-tuning LLMs. It enhances fine-tuning efficiency by reducing the number of trainable parameters while achieving performance comparable to that of fully fine-tuned models. For the model training, we employed a batch size of 8 and set the learning rate to $5e^{-5}$. The model was trained using an NVIDIA Tesla V100 32GB SXM2 GPU.
\end{itemize}
The model that showed the best performance on the validation set was then selected for further evaluation on the test set. For the cognitive distortion and suicide classification tasks, we utilized a five-fold cross-validation approach. We utilize the precision, recall, and F1-score as evaluation metrics.

For both model adaptive pretraining and finetuning stages, we use PyTorch~\cite{paszke2019pytorch} as our implementation framework. 
All the pretrained models were employed from HuggingFace v4.28.1~\cite{wolf2020transformers}. 
Detailed configurations, source codes, and our pretrained language model are made public available via: \url{https://github.com/zwzzzQAQ/Chinese-MentalBERT}. 

\begin{table*}[!h]
\centering
\resizebox{\textwidth}{!}{
\begin{tabular}{ccccccccccc}
\hline
\multicolumn{1}{c}{Method} & \multicolumn{1}{c}{ } & \multicolumn{3}{c}{ESWECT-usual} & \multicolumn{3}{c}{ESWECT-epidemic} & \multicolumn{3}{c}{Waimai\_10k}\\
\hline
& \multicolumn{1}{c}{Masking} & F1 & P & R & F1 & P & R & F1 & P & R\\
\hline
\multicolumn{11}{c}{Baselines}\\
\hline
Word2vec-BiLSTM & - & 63.39 & 63.35 & 63.85 & 53.40 & 53.73 & 53.51 & 81.89 & 86.52 & 77.72\\
\hline
Word2vec-CNN & - & 69.73 & 70.28 & 69.50 & 62.43 & 65.86 & 60.70 & 83.94 & 87.83 & 80.93\\
\hline
BERT & Random & 74.76 & 73.70 & 76.65 & 64.12 & 64.11 & 64.49 & 87.58 & \textbf{90.89} & 84.50\\
\hline
Chinese-BERT-wwm-ext & Random & 74.85 & 73.93 & 76.77 & 63.82 & 66.77 & 62.37 & 87.44 & 87.23 & 87.65\\
\hline
DKPLM-financial & Random & 74.51 & 74.41 & 74.79 & 63.67 & 64.02 & 63.92 & 86.65 & 86.86 & 86.44\\
\hline
DKPLM-medical & Random & 74.36 & 74.35 & 74.48 & 62.05 & 61.86 & 63.09 & 85.25 & 87.31 & 83.29\\
\hline
Llama2-Chinese-Chat  & - & 75.51 & 75.45 & 75.82 & 58.01 & 63.09 & 57.98 & 85.42 & 87.43 & 83.50\\
\hline
Chinese-Alpaca-2 & - & 76.04 & \textbf{76.81} & 75.42 & 61.22 & 67.32 & 60.49 & 87.55 & 86.37 & 88.75\\
\hline
\multicolumn{11}{c}{Our method}\\ 
\hline
Chinese MentalBERT & Random & 76.13 & 75.08 & \textbf{77.54} & 66.48 & \textbf{69.90} & 64.34 & 87.94 & 88.48 & 87.41\\
\hline
Chinese MentalBERT & Guided & \textbf{76.74} & 76.69 & 77.31 & \textbf{67.77} & 69.61 & \textbf{66.48} & \textbf{89.70} & 89.81 & \textbf{89.59}\\
\hline
\end{tabular}
}
\caption{Model performances on three sentiment analysis tasks. Evaluation metrics including precision (P), recall (R), and F1-score (F1) are reported as macro averages for the ESWECT-usual and ESWECT-epidemic datasets, and as binary averages for the Waimai\_10k dataset.}
\label{tab:result_emotion}
\end{table*}

\begin{table*}[!h]
\centering
\resizebox{\textwidth}{!}{
\begin{tabular}{ccccccccccc}
\hline
\multicolumn{1}{c}{Method} & \multicolumn{1}{c}{ } & \multicolumn{3}{c}{Cognitive} & \multicolumn{3}{c}{Suicide} & \multicolumn{3}{c}{C2D2}\\
\hline
& \multicolumn{1}{c}{Masking} & F1 & P & R & F1 & P & R & F1 & P & R\\
\hline
\multicolumn{11}{c}{Baselines}\\
\hline
Word2vec-BiLSTM & - & 57.89 & 70.60 & 49.06 & 72.54 & 73.52 & 72.72 & 52.06 & 52.60 & 51.93\\
\hline
Word2vec-CNN & - & 68.64 & 81.75 & 59.15 & 82.40 & 83.06 & 81.74 & 57.86 & 59.35 & 58.07\\
\hline
BERT & Random & 73.06 & \textbf{82.80} & 65.37 & 83.46 & 82.81 & 84.12 & 63.35 & 63.53 & 63.74\\
\hline
Chinese-BERT-wwm & Random & 72.66 & 80.27 & 66.37 & 84.39 & 76.28 & \textbf{94.44} & 63.66 & 64.18 & 64.25 \\
\hline
DKPLM-financial & Random & 71.11 & 79.41 & 64.38 & 83.66 & \textbf{84.00} & 83.33 & 59.14 & 60.01 & 59.41\\
\hline
DKPLM-medical & Random & 72.55 & 79.11 & 66.99 & 83.33 & 79.71 & 87.30 & 60.73 & 61.41 & 60.64\\
\hline
Llama2-Chinese-Chat & - & 65.96 & 66.29 & 65.63 & 73.09 & 62.86 & 87.30 & 57.73 & 59.81 & 57.66\\
\hline
Chinese-Alpaca-2 & - & 69.03 & 71.47 & 66.75 & 84.33 & 79.58 & 89.68 & 60.51 & 61.19 & 61.44\\
\hline
\multicolumn{11}{c}{Our method}\\ 
\hline
Chinese MentalBERT & Random & 73.30 & 81.65 & 66.50 & 85.71 & 79.59 & 92.85 & 65.30 & 65.11 & 65.94\\
\hline
Chinese MentalBERT & Guided & \textbf{74.75} & 79.88 & \textbf{70.23} & \textbf{86.15} & 83.58 & 88.88 & \textbf{65.67} & \textbf{65.66} & \textbf{66.00}\\
\hline
\end{tabular}
}
\caption{Model performance on three mental health related tasks including two cognitive distortions classification tasks (``Cognitive'' and ``C2D2'') and a suicide detection tasks. Evaluation metrics including precision (P), recall (R), and F1-score (F1). We reported the micro averages for the ``Cognitive'' task, macro averages for the ``C2D2'' task, and binary averages for the suicide detection task.}
\label{tab:result_suicide_cognitive}
\end{table*}

\subsection{Comparison methods}
We selected some representative models for performance comparison. The methods we compared can be broadly categorized into three main groups: Word2Vec-based methods (Word2Vec-BiLSTM, Word2Vec-CNN), pre-trained language models (BERT, Chinese-BERT-wwm-ext, DKPLM-financial, and DKPLM-medical), and LLMs (Llama2-Chinese-Chat and Chinese-Alpaca-2). Although LLMs are also PLMs, we list them separately for discussion because they refer to larger-scale PLMs. Our method achieved the best performance across all six datasets.

\paragraph{Word2Vec based DNNs} 
In our experiments, we built two deep neural networks using word2vec: Word2vec-CNN and Word2vec-BiLSTM. We utilized 300-dimensional word2vec word embeddings to represent the input text. 
\begin{itemize}
    \item Word2vec-CNN: this model employs two layers of 1D-CNN for text feature extraction. The first layer transforms the 300-dimensional input into a 100-dimensional output with a kernel size of 1, while the second layer uses a kernel size of 2 to capture features at different scales. The process concludes with a fully connected layer for classification. 
    \item Word2vec-BiLSTM: this model uses a single-layer BiLSTM to capture both forward and backward dependencies in the text, with input and hidden state sizes of 300 and 100, respectively. The output from the LSTM layer is averaged pooled to extract the overall features of the sequence, followed by a fully connected layer for classification.
\end{itemize}

\paragraph{BERT and whole word masking BERT} BERT is a transformer-based pretrained language model~\cite{devlin2018bert}. We utilize the Chinese pretrained BERT with a fully connected layer as a classifier. The Chinese-BERT-wwm-ext model~\cite{chinese_bert_cui2021pre}, designed for processing Chinese text, enhances the understanding of the nuanced aspects of the Chinese language through Whole Word Masking (WWM) implementation.

\paragraph{DKPLMs} 
DKPLM is a knowledge-enhanced PLM~\cite{zhang2022dkplm}. We employed two variant models for experiments, specifically DKPLM-financial and DKPLM-medical. The DKPLM-financial model is pre-trained in the financial domain and exhibits adeptness in deciphering intricate financial terms and context. The model excels in applications like sentiment analysis and market trend prediction within the financial sector. DKPLM-medical is pre-trained on an extensive collection of medical texts and proficiently comprehends medical terminology and patient narratives. The demonstrated proficiency guarantees enhanced performance in tasks such as clinical information extraction and medical literature analysis.

\paragraph{Chinese LLMs}
We selected two large-scale Chinese language models for comparison: Llama2-Chinese-Chat\footnote{\url{https://huggingface.co/FlagAlpha/Llama2-Chinese-7b-Chat}} and Chinese-Alpaca-2\footnote{\url{https://huggingface.co/hfl/chinese-alpaca-2-7b}}, each with a model size of 7 billion parameters. Both models, derived from existing large-scale frameworks, have undergone additional fine-tuning and training to enhance their processing capabilities for Chinese. 
\begin{itemize}
    \item Llama2-Chinese-Chat: this model is an adaptation of the Llama2-Chat, specifically optimized for the Chinese language. Llama2-Chat was developed through supervised fine-tuning of the Llama2 model~\cite{touvron2023llama} and boasts significant dialogue processing capabilities. This model was further enhanced by fine-tuning with a Chinese instruction dataset, significantly boosting its performance in Chinese conversational contexts.
    \item Chinese-Alpaca-2: this model was built on the LlaMA-2 framework, utilizes a 120GB Chinese corpus for incremental training to improve understanding and generation of Chinese text~\cite{cui2023efficient}. Furthermore, employing 5 million units of Chinese instructional data for precise adjustments culminated in the more refined Chinese-Alpaca-2-7B model. 
\end{itemize}

\section{Results}
We compared the performance of our proposed method, which includes two masking mechanisms (random masking and guided masking), with eight other models across six open datasets.
The experimental results can be seen in Tables~\ref{tab:result_emotion} and~\ref{tab:result_suicide_cognitive}.

For a detailed analysis, Word2Vec-CNN outperformed Word2Vec-BiLSTM in all tasks, although Word2Vec-based methods were generally have lower performance on all six datasets. Especially, On the ``C2D2'' data, their performance was 7.81\% lower in F1-score compared to our model, highlighting their limitations.
Among the four pre-trained model-based methods (BERT, Chinese-BERT-wwm-ext, DKPLM-financial and DKPLM-medical), performance differences across the six tasks were not great. In the best-performing ``Cognitive'' task, the best model BERT still had an 1.69\% point lower F1-score than our model. Also, in the ``ESWECT-epidemic'' task, the best-performing BERT model had an F1-score 3.65\% lower than our model.
For LLMs, Chinese-Alpaca-2 outperformed Llama2-Chinese-Chat in all tasks. However, LLMs generally fell short compared to pre-trained model-based methods, except for the ``ESWECT-usual'' task, where Chinese-Alpaca-2 outperformed Chinese-BERT-wwm-ext by 1.19\% points in F1-score. This indicates that traditional deep learning solutions still have advantages in domain-specific tasks. Also fine-tuning LLMs can be challenging due to their large model sizes for model training.
The superior performance of our models in all tasks is attributed to our domain-adaptive training and guided masking mechanism in mental health domain. The performance of our model with random masking was lower than with the guided masking training approach in all tasks, especially in the Waimai\_10k task, where there was a 1.76\% point F1-score gap. This shows that the good performance of our method is not only due to the collection of large datasets, but also benefits from the training mechanism based on lexicon guided masking.

Among the six tasks, we consider ``Cognitive'' and ``Suicide'' to be the best for validating model performance due to having the highest inter-method standard deviations (SD), which are 4.72 and 4.66, respectively. The high SDs indicate that the differences between methods are meaningful, suggesting that performance on these tasks does not saturate too quickly and that the model's performance can effectively highlight the differences between methods~\cite{isensee2024nnu}. The best performance achieved by our model in these two tasks reconfirms the reliability of our approach.

\begin{CJK*}{UTF8}{gbsn} 
\begin{table*}[!htbp] 
\centering
\footnotesize
\begin{tabular}{|l|l|l|l|} 
\hline
\multirow{2}{*}{\textbf{Sentence}}        & \multicolumn{3}{c|}{\textbf{Masked words prediction}}                                                \\ 
\cline{2-4}
                                          & \textbf{Chinese-BERT-wwm-ext}     & \textbf{Ours-Random}         & \textbf{Ours-Guided}          \\ 
\hline
Chinese: 经常责怪自己                           & 告诉 (tell)                         & 折磨 (torture)                   & 折磨 (torture)                    \\
{[}mask] Chinese: 经常[MASK][MASK]自己        & 提问 (question)                     & 怀疑 (doubt)                     & 怀疑 (doubt)                      \\ 
\cline{1-1}
English: Often blame myself               & 反问 (counter-question)             & 伤害 (hurt)                      & 压抑 (depress)                    \\
{[}mask] English: Often [MASK] myself     & 暗笑 (chuckle)                      & 提醒 (warn)                      & 伤残 (disable)                    \\ 
\hhline{|====|}
Chinese: 呼吸有困难                            & \multirow{2}{*}{急性 (acute)}       & \multirow{2}{*}{问题 (question)} & \multirow{2}{*}{困难 (trouble)}   \\
{[}mask] Chinese: 呼吸有[MASK][MASK]         &                                   &                                &                                 \\ 
\cline{1-1}
English: Having trouble breathing         & \multirow{2}{*}{气性 (temperament)} & \multirow{2}{*}{限制 (limit)}    & \multirow{2}{*}{压力 (pressure)}  \\
{[}mask] English: Having [MASK] breathing &                                   &                                &                                 \\ 
\hhline{|====|}
Chinese: 想到死亡的事                           & \multirow{2}{*}{以前 (before)}      & \multirow{2}{*}{以前 (before)}   & \multirow{2}{*}{难过 (saddness)}  \\
{[}mask] Chinese:~想到[MASK][MASK]的事        &                                   &                                &                                 \\ 
\cline{1-1}
English: Thinking about death             & \multirow{2}{*}{过去 (past)}        & \multirow{2}{*}{好多 (a lot)}    & \multirow{2}{*}{伤心 (grief)}     \\
{[}mask] English: Thinking about [MASK]   &                                   &                                &                                 \\
\hline
\end{tabular}
\caption{Comparative analysis of masked word prediction by three pre-trained models. This table presents the original Chinese sentences, the sentences with masked words ([mask] Chinese), and their English translations (English, [mask] English), where [MASK] indicates the masked word's position. It includes predictions from three models: the basic PLM Chinese-BERT-wwm-ext and our proposed models with two different masking mechanisms (Random and Guided), alongside the predicted Chinese words and their English translations.}
\label{tab:result_predictMask}
\end{table*}
\end{CJK*} 

\section{Qualitative comparison}
We conducted a qualitative analysis to explore the behavior and tendencies of language models by predicting masked words, using questions from the Symptom Checklist-90 (SCL-90 scale)~\cite{lr1973scl} as our experimental basis. The SCL-90 scale, a widely recognized 90-item tool for evaluating mental health, assesses nine primary psychiatric symptoms and psychological distress. In this study, specific keywords in each question of the SCL-90 scale were obscured, and the predictions made by Chinese-BERT-wwm-ext and Chinese MentalBERT—models trained through random or guided masking mechanisms were analyzed. Table~\ref{tab:result_predictMask} presents examples of these sentences and the corresponding predictions for the masked words.

\begin{CJK*}{UTF8}{gbsn}
The examples shown in the table reveal that the Chinese-BERT-wwm-ext model typically generates more general and less emotionally charged predictions compared to the proposed models. For instance, when faced with sentences expressing self-blame or difficulty breathing, the basic model opts for neutral words like ``告诉'' (tell) and ``急性'' (acute), which lack the emotional depth present in the context. In contrast, the proposed models, developed with a focus on psychological or emotional states, consistently select words that better capture the negative emotions or psychological nuances implied in the sentences, such as ``折磨'' (torture) and ``困难'' (trouble).

Comparing the two masking mechanisms (Random and Guided), highlights their different tendencies to word prediction. While both are inclined towards psychological and emotional expressions, the model trained with guided mechanism exhibits a clearer focus on accurately capturing the emotional context of the sentences. For example, in predicting masked words related to thinking about death, the guided model predicted words that directly relate to emotional states like ``难过'' (sadness) and ``伤心'' (grief), demonstrating its enhanced sensitivity towards psychological vocabularies. This suggests that the guided training mechanism employed in the model improves its ability to predict emotionally relevant context.
\end{CJK*}

\section{Conclusion}
In this paper, we present Chinese MentalBERT, the first adaptive pre-trained language model for the Chinese mental health domain to the best of our knowledge. The model features a simple yet effective domain adaptive framework, and experiments have shown its strong performance in Chinese psychology-related tasks. The domain lexicon-guided masking mechanism used in this study can adjust the model's tendency to enhance the performance of downstream tasks. Our pre-trained model is publicly available to support the advancement of this field.

In future research, we plan to validate our model across a broader range of data types and tasks, including analyzing mental health interview data and summarizing psychology-related content. Additionally, we aim to explore the use of diverse lexical resources tailored to specific tasks and employ clinical data to develop more specialized language models.

\section*{Limitations}
In qualitative comparison, Chinese MentalBERT shows a greater inclination to predict negative emotional words, whereas Chinese-BERT-wwm-ext in the general field produces more random predictions. We hypothesize this may be attributed to the guided masking mechanism, indicating its effectiveness in adjusting the model's tendencies. These tendencies can influence the model's attention on data to benefit specific tasks. The relationship between pretrained model tendencies and downstream task performance still needs to be explored.

Access to clinical textual data is constrained by the sensitivity of personal information and the stringent confidentiality requirements, leading to data scarcity in this field. Considering this challenge, this study utilized the Weibo corpus as the primary domain adaptation training resource. However, this approach may predispose the model to learn general Weibo language features over those specific to clinical reports, potentially limiting its applicability and effectiveness in clinical contexts. This emphasizes the importance of tailoring training resources to match the specific needs of the clinical environment. A limitation of the current study is the absence of a specific dataset for the clinical psychiatric domain, which lacks a comprehensive evaluation of the Chinese MentalBERT model's performance in clinical psychiatric diagnosis tasks. This underscores the necessity for future research to acquire datasets in pertinent fields and investigate the model's potential in particular clinical scenarios.

\section*{Ethics statement}
In order to mitigate the risk of disclosing personal information, we anonymize and de-identify the data to the greatest extent possible during processing and analysis. We guarantee that the research outcomes do not include any information that could directly or indirectly identify individuals. Due to data privacy concerns, the pretraining corpus will not be made available to the public. However, to further the development of the mental health field in China, we have made the pre-trained model and code public accessible to the community. It is worth noting that datasets for downstream tasks might contain biases from social media data, including gender, age, or sexual orientation profiles, potentially leading to the incorrect labeling of individuals as having a mental disorder. We emphasize that the experimentation with and utilization of these data are strictly confined to research and analysis purposes, and any misuse or mishandling of the information is expressly forbidden.

\section*{Acknowledgements}
This work was supported by grants from the National Natural Science Foundation of China (grant numbers: 72174152 and 82071546), Fundamental Research Funds for the Central Universities (grant numbers: 2042022kf1218; 2042022kf1037), the Young Top-notch Talent Cultivation Program of Hubei Province.
Guanghui Fu is supported by a Chinese Government Scholarship provided by the China Scholarship Council (CSC).

\bibliography{ref}

\begin{thebibliography}{44}
\expandafter\ifx\csname natexlab\endcsname\relax\def\natexlab#1{#1}\fi

\bibitem[{Ambalavanan et~al.(2019)Ambalavanan, Jagtap, Adhya, and Devarakonda}]{ambalavanan2019using}
Ashwin~Karthik Ambalavanan, Pranjali~Dileep Jagtap, Soumya Adhya, and Murthy Devarakonda. 2019.
\newblock Using contextual representations for suicide risk assessment from internet forums.
\newblock In \emph{Proceedings of the sixth workshop on computational linguistics and clinical psychology}, pages 172--176.

\bibitem[{Aragon et~al.(2023)Aragon, Monroy, Gonzalez, Losada, and Montes}]{aragon2023disorbert}
Mario Aragon, Adri{\'a}n Pastor~L{\'o}pez Monroy, Luis Gonzalez, David~E Losada, and Manuel Montes. 2023.
\newblock {DisorBERT}: A double domain adaptation model for detecting signs of mental disorders in social media.
\newblock In \emph{Proceedings of the 61st Annual Meeting of the Association for Computational Linguistics (Volume 1: Long Papers)}, pages 15305--15318.

\bibitem[{Burns(1981)}]{burns1981feeling}
David~D Burns. 1981.
\newblock \emph{Feeling good}.
\newblock Signet Book.

\bibitem[{Cai et~al.(2023)Cai, Wang, Ye, Jin, and Gao}]{dataset_cai2023depression}
Yicheng Cai, Haizhou Wang, Huali Ye, Yanwen Jin, and Wei Gao. 2023.
\newblock Depression detection on online social network with multivariate time series feature of user depressive symptoms.
\newblock \emph{Expert Systems with Applications}, 217:119538.

\bibitem[{Calvo et~al.(2017)Calvo, Milne, Hussain, and Christensen}]{calvo2017natural}
Rafael~A Calvo, David~N Milne, M~Sazzad Hussain, and Helen Christensen. 2017.
\newblock Natural language processing in mental health applications using non-clinical texts.
\newblock \emph{Natural Language Engineering}, 23(5):649--685.

\bibitem[{Chalkidis et~al.(2020)Chalkidis, Fergadiotis, Malakasiotis, Aletras, and Androutsopoulos}]{chalkidis2020legal}
Ilias Chalkidis, Manos Fergadiotis, Prodromos Malakasiotis, Nikolaos Aletras, and Ion Androutsopoulos. 2020.
\newblock \href {https://doi.org/10.18653/v1/2020.findings-emnlp.261} {{LEGAL}-{BERT}: The muppets straight out of law school}.
\newblock In \emph{Findings of the Association for Computational Linguistics: EMNLP 2020}, pages 2898--2904, Online. Association for Computational Linguistics.

\bibitem[{Coppersmith et~al.(2018)Coppersmith, Leary, Crutchley, and Fine}]{coppersmith2018natural}
Glen Coppersmith, Ryan Leary, Patrick Crutchley, and Alex Fine. 2018.
\newblock Natural language processing of social media as screening for suicide risk.
\newblock \emph{Biomedical informatics insights}, 10:1178222618792860.

\bibitem[{Cui et~al.(2021)Cui, Che, Liu, Qin, and Yang}]{chinese_bert_cui2021pre}
Yiming Cui, Wanxiang Che, Ting Liu, Bing Qin, and Ziqing Yang. 2021.
\newblock Pre-training with whole word masking for {Chinese BERT}.
\newblock \emph{IEEE/ACM Transactions on Audio, Speech, and Language Processing}, 29:3504--3514.

\bibitem[{Cui et~al.(2023)Cui, Yang, and Yao}]{cui2023efficient}
Yiming Cui, Ziqing Yang, and Xin Yao. 2023.
\newblock Efficient and effective text encoding for {Chinese} llama and alpaca.
\newblock \emph{arXiv preprint arXiv:2304.08177}.

\bibitem[{De~Choudhury et~al.(2013)De~Choudhury, Gamon, Counts, and Horvitz}]{de2013predicting}
Munmun De~Choudhury, Michael Gamon, Scott Counts, and Eric Horvitz. 2013.
\newblock Predicting depression via social media.
\newblock In \emph{Proceedings of the international AAAI conference on web and social media}, volume~7, pages 128--137.

\bibitem[{Depression(2024)}]{chaohua}
Depression. 2024.
\newblock \href {https://weibo.com/p/100808f86f9e10c1d3bdefe430d95f95388c90/super_index} {Sina weibo "{Depression}" chaohua (super topic)}.

\bibitem[{Devlin et~al.(2018)Devlin, Chang, Lee, and Toutanova}]{devlin2018bert}
Jacob Devlin, Ming-Wei Chang, Kenton Lee, and Kristina Toutanova. 2018.
\newblock {BERT}: Pre-training of deep bidirectional transformers for language understanding.
\newblock \emph{arXiv preprint arXiv:1810.04805}.

\bibitem[{Fu et~al.(2021)Fu, Song, Li, Ma, Chen, Wang, Yang, and Huang}]{fu2021distant}
Guanghui Fu, Changwei Song, Jianqiang Li, Yue Ma, Pan Chen, Ruiqian Wang, Bing~Xiang Yang, and Zhisheng Huang. 2021.
\newblock Distant supervision for mental health management in social media: suicide risk classification system development study.
\newblock \emph{Journal of medical internet research}, 23(8):e26119.

\bibitem[{Gururangan et~al.(2020)Gururangan, Marasovi{\'c}, Swayamdipta, Lo, Beltagy, Downey, and Smith}]{domain_adaptative_gururangan2020don}
Suchin Gururangan, Ana Marasovi{\'c}, Swabha Swayamdipta, Kyle Lo, Iz~Beltagy, Doug Downey, and Noah~A. Smith. 2020.
\newblock \href {https://doi.org/10.18653/v1/2020.acl-main.740} {Don{'}t stop pretraining: Adapt language models to domains and tasks}.
\newblock In \emph{Proceedings of the 58th Annual Meeting of the Association for Computational Linguistics}, pages 8342--8360, Online. Association for Computational Linguistics.

\bibitem[{He et~al.(2023)He, Fu, Yu, Wang, Li, Zhao, Song, Qi, Luo, Zou et~al.}]{he2023towards}
Tianyu He, Guanghui Fu, Yijing Yu, Fan Wang, Jianqiang Li, Qing Zhao, Changwei Song, Hongzhi Qi, Dan Luo, Huijing Zou, et~al. 2023.
\newblock Towards a psychological generalist {AI}: A survey of current applications of large language models and future prospects.
\newblock \emph{arXiv preprint arXiv:2312.04578}.

\bibitem[{Hu et~al.(2021)Hu, Shen, Wallis, Allen-Zhu, Li, Wang, Wang, and Chen}]{hu2021lora}
Edward~J Hu, Yelong Shen, Phillip Wallis, Zeyuan Allen-Zhu, Yuanzhi Li, Shean Wang, Lu~Wang, and Weizhu Chen. 2021.
\newblock {LoRA}: Low-rank adaptation of large language models.
\newblock \emph{arXiv preprint arXiv:2106.09685}.

\bibitem[{Huang et~al.(2019)Huang, Wang, Wang, Liu, Yu, Yan, Yu, Kou, Xu, Lu et~al.}]{huang2019prevalence}
Yueqin Huang, YU~Wang, Hong Wang, Zhaorui Liu, Xin Yu, Jie Yan, Yaqin Yu, Changgui Kou, Xiufeng Xu, Jin Lu, et~al. 2019.
\newblock Prevalence of mental disorders in {China}: a cross-sectional epidemiological study.
\newblock \emph{The Lancet Psychiatry}, 6(3):211--224.

\bibitem[{Isensee et~al.(2024)Isensee, Wald, Ulrich, Baumgartner, Roy, Maier-Hein, and Jaeger}]{isensee2024nnu}
Fabian Isensee, Tassilo Wald, Constantin Ulrich, Michael Baumgartner, Saikat Roy, Klaus Maier-Hein, and Paul~F Jaeger. 2024.
\newblock {nnU-Net} revisited: A call for rigorous validation in {3D} medical image segmentation.
\newblock \emph{arXiv preprint arXiv:2404.09556}.

\bibitem[{Ji et~al.(2022)Ji, Zhang, Ansari, Fu, Tiwari, and Cambria}]{ji2021mentalbert}
Shaoxiong Ji, Tianlin Zhang, Luna Ansari, Jie Fu, Prayag Tiwari, and Erik Cambria. 2022.
\newblock \href {https://aclanthology.org/2022.lrec-1.778} {{M}ental{BERT}: Publicly available pretrained language models for mental healthcare}.
\newblock In \emph{Proceedings of the Thirteenth Language Resources and Evaluation Conference}, pages 7184--7190, Marseille, France. European Language Resources Association.

\bibitem[{Jin et~al.(2022)Jin, Zhang, Zhu, Xiao, Li, Wei, Arnold, and Ren}]{jin2021lifelong}
Xisen Jin, Dejiao Zhang, Henghui Zhu, Wei Xiao, Shang-Wen Li, Xiaokai Wei, Andrew Arnold, and Xiang Ren. 2022.
\newblock \href {https://doi.org/10.18653/v1/2022.bigscience-1.1} {Lifelong pretraining: Continually adapting language models to emerging corpora}.
\newblock In \emph{Proceedings of BigScience Episode {\#}5 -- Workshop on Challenges {\&} Perspectives in Creating Large Language Models}, pages 1--16, virtual+Dublin. Association for Computational Linguistics.

\bibitem[{Keles et~al.(2020)Keles, McCrae, and Grealish}]{keles2020systematic}
Betul Keles, Niall McCrae, and Annmarie Grealish. 2020.
\newblock A systematic review: the influence of social media on depression, anxiety and psychological distress in adolescents.
\newblock \emph{International journal of adolescence and youth}, 25(1):79--93.

\bibitem[{Kingma and Ba(2014)}]{kingma2014adam}
Diederik~P Kingma and Jimmy Ba. 2014.
\newblock Adam: A method for stochastic optimization.
\newblock \emph{arXiv preprint arXiv:1412.6980}.

\bibitem[{Koroteev(2021)}]{koroteev2021bert}
MV~Koroteev. 2021.
\newblock {BERT}: a review of applications in natural language processing and understanding.
\newblock \emph{arXiv preprint arXiv:2103.11943}.

\bibitem[{Lee et~al.(2020)Lee, Yoon, Kim, Kim, Kim, So, and Kang}]{lee2020biobert}
Jinhyuk Lee, Wonjin Yoon, Sungdong Kim, Donghyeon Kim, Sunkyu Kim, Chan~Ho So, and Jaewoo Kang. 2020.
\newblock {BioBERT}: a pre-trained biomedical language representation model for biomedical text mining.
\newblock \emph{Bioinformatics}, 36(4):1234--1240.

\bibitem[{Li et~al.(2020)Li, Li, Huang, Hou et~al.}]{lexicon_li2020automatic}
Genghao Li, Bing Li, Langlin Huang, Sibing Hou, et~al. 2020.
\newblock Automatic construction of a depression-domain lexicon based on microblogs: text mining study.
\newblock \emph{JMIR medical informatics}, 8(6):e17650.

\bibitem[{LR et~al.(1973)LR, RS, and Covi}]{lr1973scl}
Derogatis LR, Lipman RS, and L~Covi. 1973.
\newblock {SCL}-90: an outpatient psychiatric rating scale-preliminary report.
\newblock \emph{Psychopharmacol Bull}, 9:13--28.

\bibitem[{Matero et~al.(2019)Matero, Idnani, Son, Giorgi, Vu, Zamani, Limbachiya, Guntuku, and Schwartz}]{matero2019suicide}
Matthew Matero, Akash Idnani, Youngseo Son, Salvatore Giorgi, Huy Vu, Mohammad Zamani, Parth Limbachiya, Sharath~Chandra Guntuku, and H~Andrew Schwartz. 2019.
\newblock Suicide risk assessment with multi-level dual-context language and {BERT}.
\newblock In \emph{Proceedings of the sixth workshop on computational linguistics and clinical psychology}, pages 39--44.

\bibitem[{Organization et~al.(2014)}]{world2014preventing}
World~Health Organization et~al. 2014.
\newblock \emph{Preventing suicide: A global imperative}.
\newblock World Health Organization.

\bibitem[{Organization et~al.(2023)}]{world2023depressive}
World~Health Organization et~al. 2023.
\newblock Depressive disorder (depression); 2023.
\newblock \emph{Accessed: April}, 4.

\bibitem[{Paszke et~al.(2019)Paszke, Gross, Massa, Lerer, Bradbury, Chanan, Killeen, Lin, Gimelshein, Antiga et~al.}]{paszke2019pytorch}
Adam Paszke, Sam Gross, Francisco Massa, Adam Lerer, James Bradbury, Gregory Chanan, Trevor Killeen, Zeming Lin, Natalia Gimelshein, Luca Antiga, et~al. 2019.
\newblock {PyTorch}: An imperative style, high-performance deep learning library.
\newblock \emph{Advances in neural information processing systems}, 32.

\bibitem[{Primack et~al.(2017)Primack, Shensa, Sidani, Whaite, yi~Lin, Rosen, Colditz, Radovic, and Miller}]{primack2017social}
Brian~A Primack, Ariel Shensa, Jaime~E Sidani, Erin~O Whaite, Liu yi~Lin, Daniel Rosen, Jason~B Colditz, Ana Radovic, and Elizabeth Miller. 2017.
\newblock Social media use and perceived social isolation among young adults in the {US}.
\newblock \emph{American journal of preventive medicine}, 53(1):1--8.

\bibitem[{Qi et~al.(2023)Qi, Zhao, Song, Zhai, Luo, Liu, Yu, Wang, Zou, Yang et~al.}]{qi2023evaluating}
Hongzhi Qi, Qing Zhao, Changwei Song, Wei Zhai, Dan Luo, Shuo Liu, Yi~Jing Yu, Fan Wang, Huijing Zou, Bing~Xiang Yang, et~al. 2023.
\newblock Evaluating the efficacy of supervised learning vs large language models for identifying cognitive distortions and suicidal risks in chinese social media.
\newblock \emph{arXiv preprint arXiv:2309.03564}.

\bibitem[{Robinson et~al.(2016)Robinson, Cox, Bailey, Hetrick, Rodrigues, Fisher, and Herrman}]{robinson2016social}
Jo~Robinson, Georgina Cox, Eleanor Bailey, Sarah Hetrick, Maria Rodrigues, Steve Fisher, and Helen Herrman. 2016.
\newblock Social media and suicide prevention: a systematic review.
\newblock \emph{Early intervention in psychiatry}, 10(2):103--121.

\bibitem[{R{\"u}sch et~al.(2005)R{\"u}sch, Angermeyer, and Corrigan}]{rusch2005mental}
Nicolas R{\"u}sch, Matthias~C Angermeyer, and Patrick~W Corrigan. 2005.
\newblock Mental illness stigma: Concepts, consequences, and initiatives to reduce stigma.
\newblock \emph{European psychiatry}, 20(8):529--539.

\bibitem[{Shamshiri et~al.(2024)Shamshiri, Ryu, and Park}]{shamshiri2024text}
Alireza Shamshiri, Kyeong~Rok Ryu, and June~Young Park. 2024.
\newblock Text mining and natural language processing in construction.
\newblock \emph{Automation in Construction}, 158:105200.

\bibitem[{Tian et~al.(2020)Tian, Gao, Xiao, Liu, He, Wu, Wang, and Wu}]{tian2020skep}
Hao Tian, Can Gao, Xinyan Xiao, Hao Liu, Bolei He, Hua Wu, Haifeng Wang, and Feng Wu. 2020.
\newblock \href {https://doi.org/10.18653/v1/2020.acl-main.374} {{SKEP}: Sentiment knowledge enhanced pre-training for sentiment analysis}.
\newblock In \emph{Proceedings of the 58th Annual Meeting of the Association for Computational Linguistics}, pages 4067--4076, Online. Association for Computational Linguistics.

\bibitem[{Touvron et~al.(2023)Touvron, Martin, Stone, Albert, Almahairi, Babaei, Bashlykov, Batra, Bhargava, Bhosale et~al.}]{touvron2023llama}
Hugo Touvron, Louis Martin, Kevin Stone, Peter Albert, Amjad Almahairi, Yasmine Babaei, Nikolay Bashlykov, Soumya Batra, Prajjwal Bhargava, Shruti Bhosale, et~al. 2023.
\newblock Llama 2: Open foundation and fine-tuned chat models.
\newblock \emph{arXiv preprint arXiv:2307.09288}.

\bibitem[{Wang et~al.(2023)Wang, Deng, Zhao, and Qin}]{wang2023c2d2}
Bichen Wang, Pengfei Deng, Yanyan Zhao, and Bing Qin. 2023.
\newblock {C2D2} dataset: A resource for the cognitive distortion analysis and its impact on mental health.
\newblock In \emph{The 2023 Conference on Empirical Methods in Natural Language Processing}.

\bibitem[{Wang et~al.(2019)Wang, Chen, Li, Li, Zhou, Zheng, Zhang, and Tang}]{wang2019assessing}
Xiaofeng Wang, Shuai Chen, Tao Li, Wanting Li, Yejie Zhou, Jie Zheng, Yaoyun Zhang, and Buzhou Tang. 2019.
\newblock Assessing depression risk in {Chinese} microblogs: a corpus and machine learning methods.
\newblock In \emph{2019 IEEE International Conference on Healthcare Informatics (ICHI)}, pages 1--5. IEEE.

\bibitem[{Wang et~al.(2020)Wang, Wang, Li, Zhang, and Wang}]{dataset_wang2020multitask}
Yiding Wang, Zhenyi Wang, Chenghao Li, Yilin Zhang, and Haizhou Wang. 2020.
\newblock A multitask deep learning approach for user depression detection on sina weibo.
\newblock \emph{arXiv preprint arXiv:2008.11708}.

\bibitem[{Wolf et~al.(2020)Wolf, Debut, Sanh, Chaumond, Delangue, Moi, Cistac, Rault, Louf, Funtowicz et~al.}]{wolf2020transformers}
Thomas Wolf, Lysandre Debut, Victor Sanh, Julien Chaumond, Clement Delangue, Anthony Moi, Pierric Cistac, Tim Rault, R{\'e}mi Louf, Morgan Funtowicz, et~al. 2020.
\newblock Transformers: State-of-the-art natural language processing.
\newblock In \emph{Proceedings of the 2020 conference on empirical methods in natural language processing: system demonstrations}, pages 38--45.

\bibitem[{Yang et~al.(2022)Yang, Chen, Li, Yang, Huang, Fu, Luo, Wang, Li, Wen et~al.}]{yang2022characteristics}
Bing~Xiang Yang, Pan Chen, Xin~Yi Li, Fang Yang, Zhisheng Huang, Guanghui Fu, Dan Luo, Xiao~Qin Wang, Wentian Li, Li~Wen, et~al. 2022.
\newblock Characteristics of high suicide risk messages from users of a social network—sina weibo “tree hole”.
\newblock \emph{Frontiers in psychiatry}, 13:789504.

\bibitem[{Zhang et~al.(2022)Zhang, Wang, Hu, Qiu, Tang, He, and Huang}]{zhang2022dkplm}
Taolin Zhang, Chengyu Wang, Nan Hu, Minghui Qiu, Chengguang Tang, Xiaofeng He, and Jun Huang. 2022.
\newblock {DKPLM}: decomposable knowledge-enhanced pre-trained language model for natural language understanding.
\newblock In \emph{Proceedings of the AAAI Conference on Artificial Intelligence}, volume~36, pages 11703--11711.

\bibitem[{ZouFan(2023)}]{zoufan}
ZouFan. 2023.
\newblock \href {https://www.weibo.com/xiaofan116?is_all=1} {Sina weibo "{Zoufan}" comment}.

\end{thebibliography}
\bibliographystyle{acl_natbib}



\appendix
\renewcommand\thesection{Appendix}
\renewcommand\thefigure{S\arabic{figure}} 
\setcounter{figure}{0} 
\renewcommand\thetable{S\arabic{table}} 
\setcounter{table}{0}

\section{Error analysis}
We conducted a detailed analysis of model performance across each major category of cognitive distortions in the ``Cognitive'' dataset. Given the limited number of cognitive distortion types in four of the categories within the dataset, our discussion primarily focuses on the experimental results of the other eight categories. The experimental results are detailed in Table~\ref{tab:result_error}. The results indicate that the proposed Chinese MentalBERT outperforms other models in seven of the eight primary categories of the cognitive distortion multi-label classification task, while it exhibits slightly inferior performance in one category. Addressing this discrepancy will be the primary focus of our future research endeavors. 


\begin{table}[!h]
\centering
\resizebox{1.0\linewidth}{!}{
\begin{tabular}{ccccc}
\hline
\textbf{Method} & \textbf{C1} & \textbf{C2} & \textbf{C3} & \textbf{C4} \\
\hline
BERT & 0.00 & \textbf{55.56} & 23.53 & 70.39 \\
\hline
Chinese-BERT-wwm-ext & 0.00 & 55.17 & 30.00 & 70.54 \\
\hline
Chinese MentalBERT  & \textbf{6.25} & 55.07 & \textbf{40.91} & \textbf{72.58} \\
\hline
\textbf{Method} & \textbf{C5} & \textbf{C6} & \textbf{C7} & \textbf{C8} \\
\hline
BERT & 31.58 & 62.07 & 92.50 & 65.71 \\
\hline
Chinese-BERT-wwm-ext & 18.18 & 57.14 & 92.12 & 48.28 \\
\hline
Chinese MentalBERT & \textbf{34.19} & \textbf{72.00} & \textbf{92.67} & \textbf{66.67} \\
\hline
\end{tabular}}
\caption{Performance (F1-score) comparison of three models across the eight cognitive distortion categories within the ``Cognitive'' dataset. C1: Over-generalization, C2: Mental Filter, C3: Mind Reading, C4: The Fortune Teller Error, C5: Magnification, C6: Should Statements, C7: Labeling and Mislabeling, C8: Blaming Oneself.}
\label{tab:result_error}
\end{table}

\end{document}